%% file: main.tex
\documentclass[runningheads]{llncs}

\usepackage[mobile]{eccv}

\usepackage{eccvabbrv}

\usepackage{graphicx}
\usepackage{booktabs}

\usepackage[accsupp]{axessibility}

\usepackage{hyperref}

\usepackage{orcidlink}

\begin{document}

\title{End-to-End Spatial-Temporal Transformer for Real-time 4D HOI Reconstruction} 

\titlerunning{THO for Real-time 4D HOI Reconstruction}

\author{Haoyu Zhang \and
Wei Zhai\textsuperscript{\dag} \and
Yuhang Yang \and
Yang Cao \and
Zheng-Jun Zha}

\authorrunning{Zhang et al.}
\institute{University of Science and Technology of China \\
\email{\{nianheng@mail.,wzhai056@,yyuhang@mail.,forrest@,zhazj@\}ustc.edu.cn}}

\maketitle
\let\thefootnote\relax\footnotetext{\textsuperscript{\dag} Corresponding author.}

\begin{center}
  \begin{minipage}{0.993\linewidth}
    \centering
    \includegraphics[width=0.98\linewidth]{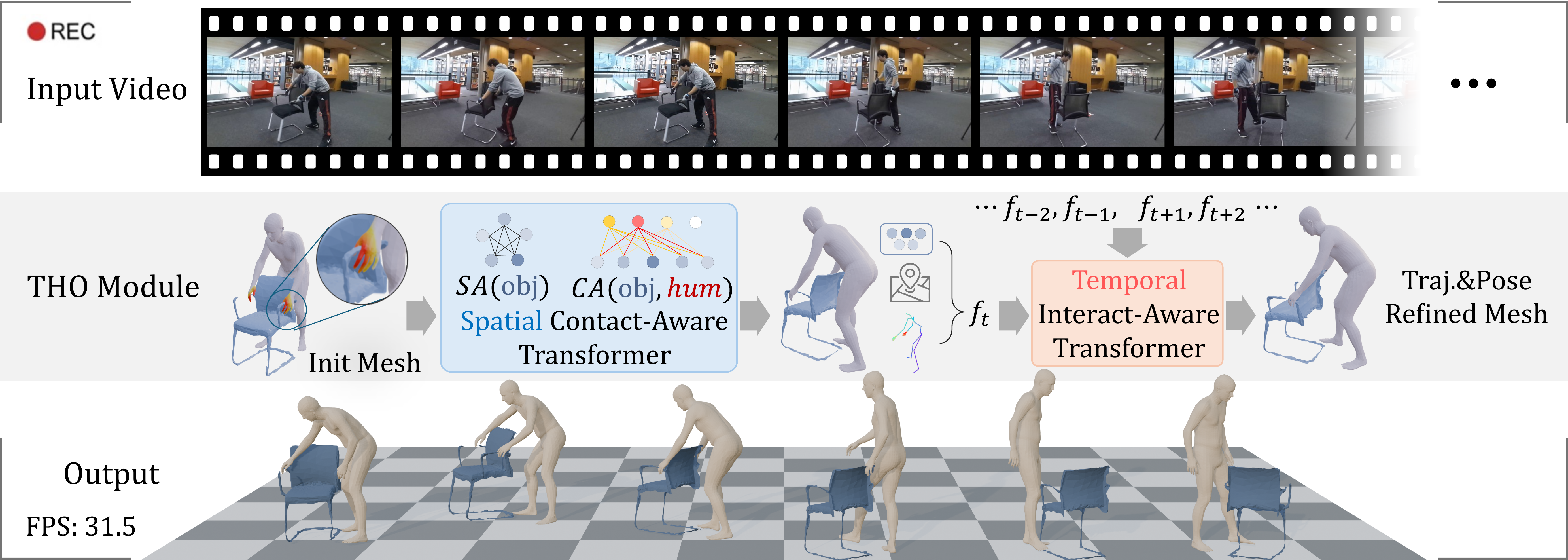}
    \captionof{figure}{\textbf{Overview.} THO enables real-time 4D HOI reconstruction (31.5 FPS) by introducing a Spatial-Temporal Transformer to capture contact and interaction dynamics.}
    \label{fig:teaser}
  \end{minipage}
\end{center}

\begin{abstract}
Monocular 4D human–object interaction (HOI) reconstruction—recovering a moving human and a manipulated object from a single RGB video—remains challenging due to depth ambiguity and frequent occlusions. Existing methods often rely on multi-stage pipelines or iterative optimization, leading to high inference latency, failing to meet real-time requirements, and susceptibility to error accumulation. To address these limitations, we propose THO, an end-to-end Spatial-Temporal Transformer that predicts human motion and coordinated object motion in a forward fashion from the given video and 3D template. THO achieves this by leveraging spatial-temporal HOI tuple priors. Spatial priors exploit contact-region proximity to infer occluded object features from human cues, while temporal priors capture cross-frame kinematic correlations to refine object representations and enforce physical coherence. Extensive experiments demonstrate that THO operates at an inference speed of 31.5 FPS on a single RTX 4090 GPU, achieving a $>600\times$ speedup over prior optimization-based methods while simultaneously improving reconstruction accuracy and temporal consistency. The project page is \href{https://nianheng.github.io/THO-project/}{here}.
\end{abstract}

\input{paragraphs/1_Introduction}
\input{paragraphs/2_Related_work}
\input{paragraphs/3_Method}
\input{paragraphs/4_Experiments}
\input{paragraphs/5_Conclusion}

\bibliographystyle{splncs04}
\bibliography{main}

\clearpage

\begin{center}
    \Large \textbf{End-to-End Spatial-Temporal Transformer for Real-time 4D HOI Reconstruction}
\end{center}

\begin{center}
    \Large \textbf{Supplementary Material}
\end{center}

This supplementary material provides comprehensive technical details to complement the main manuscript. Specifically, we elaborate on the network architecture, present the complete formulations and weights for our training objectives, and offer an in-depth analysis of the framework's limitations.

\input{paragraphs/6_details}
\input{paragraphs/7_limitations}

\end{document}

%% file: paragraphs/1_introduction.tex
\section{Introduction}
\label{sec:Introduction}

Monocular 4D human–object interaction (HOI) reconstruction seeks to recover a temporally coherent 3D human motion together with the manipulated object motion from an RGB video~\cite{bhatnagar2022behave, huang2022intercap, liu2022hoi4d}. This capability is important for applications such as AR/VR, embodied AI, and animation~\cite{wen2025open3dhoi, zhang2025interactanything, liu2025core4d}, where accurate relative motion and long-term stability matter as much as per-frame geometry. However, monocular inputs introduce severe depth ambiguity, and HOI sequences exhibit frequent mutual occlusions and rapid contact changes, making it difficult to maintain consistent reconstructions over time~\cite{xie2025cari4d, yuan2022glamr}. These challenges become especially acute when targeting real-time inference speed, which is required for interactive deployment~\cite{chen2026human3r, wang2025josh}.

Most existing solutions address monocular HOI via multi-stage pipelines with optimization-in-the-loop~\cite{xie2023vistracker, xie2024intertrack, xie2025cari4d}. Representative examples, such as VisTracker~\cite{xie2023vistracker} and the category-agnostic CARI4D~\cite{xie2025cari4d}, rely on joint optimization to satisfy image and contact evidence. While effective at enforcing constraints, such optimization-centric designs are computationally heavy, require careful initialization, and suffer from stage-wise error propagation, making interactive deployment challenging. In parallel, methods like CONTHO~\cite{nam2024contho} and HOI-TG~\cite{wang2025endtoendhoi} move toward feed-forward prediction, producing strong single-frame reconstructions without expensive test-time optimization. However, these approaches do not explicitly address temporal consistency; consequently, applying them frame-by-frame often yields jitter, drift, and inconsistent relative motion under occlusion~\cite{kocabas2020vibe, zeng2022smoothnet}. This limitation motivates a model that is both optimization-free and explicitly temporal-aware to enforce motion coherence.

To overcome these limitations, we propose THO, an end-to-end feed-forward model for real-time 4D HOI reconstruction. While such mapping-based formulations enable fast inference, they also face a fundamental challenge: sparse visual evidence caused by frequent occlusions. It inevitably introduces severe pose ambiguities—where multiple object poses may appear equally valid when critical geometric features are hidden~\cite{weng2020holistic, xie2022chore}. To resolve this ambiguity, we leverage the structured nature of human–object interactions, which provide complementary spatial and temporal cues for object reasoning. Spatially, our core insight is to exploit the physical contact between the human and the object as a strong prior. By leveraging the dual consistency of semantic correspondence and spatial proximity at contact regions, our approach utilizes unoccluded object contexts and interacting human features to effectively infer and reconstruct the hidden representations~\cite{nam2024contho, wang2025endtoendhoi}. Temporally, we recognize that human-object interactions are continuous physical processes rather than isolated frames. By explicitly modeling the kinematic correlations between human joints and the object, we leverage cross-frame interaction dynamics to correct current-frame features and enforce long-term physical coherence~\cite{goel2023humans4d, choi2021tcmr}.

Specifically, THO employs GVHMR~\cite{shen2024gvhmr} to establish human motion priors and extracts interaction-centric image patches to decouple local interactions from global translation. A 3D Vertex Encoder then initializes object geometry and constructs unified geometry-aware representations for both human and object by fusing projected DINOv3 visual features~\cite{simeoni25dinov3} with 3D coordinates. To mitigate occlusions, the Spatial Contact-Aware Transformer reconstructs occluded object regions via self-attention and injects human interaction priors via cross-attention, refining object features based on the dual consistency of semantic and spatial proximity. Subsequently, a Temporal Interact-Aware Transformer processes motion tokens fusing refined object features, human joint features, and global context embeddings, employing Rotary Positional Embeddings (RoPE)~\cite{su2021roformer} and local attention to capture continuous relative dynamics.

Experiments on the BEHAVE~\cite{bhatnagar2022behave} and InterCap~\cite{huang2022intercap} benchmarks demonstrate that THO outperforms optimization-based counterparts in both reconstruction accuracy and temporal smoothness. Specifically, on the BEHAVE dataset, our approach improves the combined Chamfer distance ($CD_{c}$) by 13.4\% and reduces object acceleration error ($Acc_{o}$) by 15.7\% over state-of-the-art methods. Crucially, THO achieves this superior reconstruction quality without computationally expensive test-time optimization, pioneering real-time 4D HOI reconstruction at an inference speed of 31.5 FPS on a single RTX 4090 GPU.

In summary, our contributions are threefold:
\begin{itemize}
\item We propose THO, a transformer-based framework that reformulates 4D HOI reconstruction from test-time optimization to an efficient feed-forward mapping paradigm, enabling fast and accurate inference.

\item We introduce the Spatial Contact-Aware Transformer and Temporal Interact-Aware Transformer, addressing the pose ambiguities and temporal inconsistencies problem in mapping-based methods that arise from sparse visual evidence under frequent occlusions.

\item Extensive experiments demonstrate that our approach achieves real-time inference speed and outperforms state-of-the-art methods in both reconstruction accuracy and temporal smoothness.
\end{itemize}

%% file: paragraphs/2_Related_work.tex
\section{Related Work}
\label{sec:related_work}

\noindent\textbf{Human Motion Reconstruction.} Monocular human motion recovery provides the kinematic foundation for HOI. Early single-frame methods have evolved into video-based approaches prioritizing temporal consistency and occlusion robustness. These techniques effectively aggregate temporal cues to smooth jitter~\cite{kocabas2020vibe,choi2021tcmr,zeng2022smoothnet,zhang2021pymaf}, using attention mechanisms to handle truncations and occlusions in dynamic scenes~\cite{kocabas2021pare,li2022cliff,tokenhmr2024}. Crucially, the field has shifted toward recovering world-grounded global trajectories~\cite{yuan2022glamr,goel2023humans4d}. State-of-the-art methods like WHAM~\cite{shin2024wham} and GVHMR~\cite{shen2024gvhmr} explicitly leverage gravity and motion priors to resolve depth ambiguities, enabling stable long-term tracking. Although advancements in high-fidelity avatar generation~\cite{ho2024sith,guo2024reloo,guo2025vid2avatarpro,yang2025sigman} and controllable human animation~\cite{li2024dispose,black2023bedlam,patel2021agora} enhance visual quality, their high computational cost precludes real-time interactive applications. Our approach builds upon these global motion priors to ground interactions in a coherent 4D space.

\noindent\textbf{3D HOI Reconstruction.} Monocular human--object spatial reasoning is challenging due to depth ambiguity. While early optimization-based attempts treated spatial constraints separately~\cite{zhang2020phosa,hasson2020photometric}, modern approaches favor joint reconstruction to enforce mutual consistency~\cite{weng2020holistic}. To improve geometric plausibility, methods have explored learning implicit representations~\cite{xie2022chore}, defining contact potential fields~\cite{yang2021cpf}, and grounding object affordances from 2D relations~\cite{yang2023grounding,yang2024lemon}. To handle visual sparsity under occlusion, Transformer-based architectures like CONTHO~\cite{nam2024contho} and HOI-TG~\cite{wang2025endtoendhoi} effectively inject contact priors to refine interacting features. Concurrently, generative models are widely adopted to reason about and hallucinate plausible interactions~\cite{li2025gir} for both hand-object~\cite{ye2023diffhoi,ye2024ghop,han2025touch}, full-body scenarios~\cite{diller2024cghoi,li2025scorehoi,xie2024procigen}, and even human reactions~\cite{yu2025hero}. Despite expanding toward open-vocabulary and in-the-wild generalization~\cite{wen2025open3dhoi,dwivedi2025interactvlm,huo2024wildhoi,cseke2025pico}, most of these methods operate on static frames. When applied to video, these frame-independent predictions often suffer from temporal jitter and lack mechanisms to maintain consistent relative motion over time.

\noindent\textbf{4D HOI Reconstruction.} Reconstructing interactions in spacetime introduces complex constraints on temporal coherence and frequent contact changes. High-fidelity tracking is achievable using multi-view setups, depth sensors, or IMUs, as demonstrated by benchmarks like BEHAVE, InterCap, HOI4D, and egocentric captures~\cite{bhatnagar2022behave,huang2022intercap,liu2022hoi4d,zhang2023neuraldome,su2021robustfusion,zhang2024hoim3,zhao2024imhoi,yang2024egochoir}. However, specialized capture equipment limits consumer-grade deployment. In the monocular RGB domain, general object trackers~\cite{wen2023bundlesdf,wen2024foundationpose} perform well on rigid motion but ignore human interaction cues. Bridging this gap, interaction-centric approaches like VisTracker~\cite{xie2023vistracker} and InterTrack~\cite{xie2024intertrack} employ optimization-in-the-loop tracking, imposing contact and visibility constraints. Although CARI4D extends this to category-agnostic reconstruction~\cite{xie2025cari4d}, it retains an optimization-heavy design for stability. While these methods produce consistent results, they suffer from slow inference speeds and initialization sensitivity. Conversely, feed-forward predictors~\cite{nam2024contho,li2025scorehoi} offer speed but fail to enforce long-term coherence. While recent leaps in video generation and evaluation~\cite{zeng2024dawn,yang2025videogen,chen2025dancetogether} excel at creating visually coherent temporal interactions, adapting these heavy architectures for real-time inference remains prohibitive.

Targeting this gap, we propose THO, a real-time, feed-forward framework. By explicitly modeling temporal dynamics and contact dependencies via attention mechanisms, our approach eliminates the need for test-time optimization while maintaining robust 4D consistency.

%% file: paragraphs/3_Method.tex
\section{Method}
\label{sec:Method}

We present THO, the first real-time framework for 4D HOI reconstruction. Our core idea is to establish a spatial-temporal interact-aware network that explicitly models frame-wise contact dependencies and temporal interact dynamics. By leveraging robust human motion priors, our method guides the human and object towards a realistic, interaction-coherent reconstruction, effectively resolving geometric ambiguities caused by frequent occlusions.

Formally, given a monocular video sequence $\mathcal{I}=\{I_t \in \mathbb{R}^{H \times W \times 5}\}_{t=1}^N$ and a rigid 3D object template, where each 5-channel frame comprises a 3-channel RGB image and its corresponding 1-channel pixel-wise human and object masks, our goal is to reconstruct the temporally coherent human mesh sequence $\{M^{h}_t\}_{t=1}^N$ and determine the object's global 6D pose trajectory, parameterized by rotation $\{\hat{R}_t\}_{t=1}^N$ and translation $\{\hat{T}_t\}_{t=1}^N$ relative to the template.

\begin{figure}[tb]
  \centering
  \includegraphics[width=\linewidth]{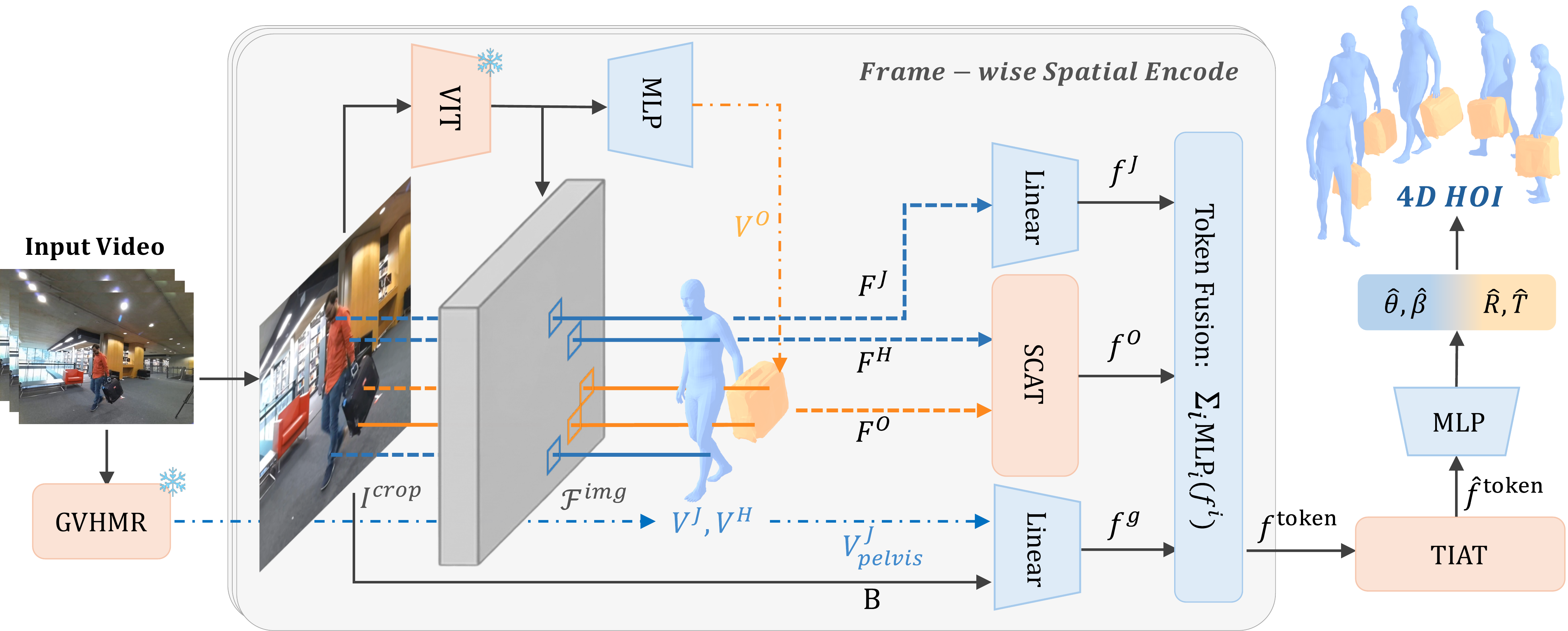}
  \caption{\textbf{Overall pipeline of THO.} From a monocular video, GVHMR extracts human priors and Interaction-Centric Crops ($\mathcal{I}^{crop}$). The 3D Vertex Encoder generates unified embeddings ($F^J, F^H, F^O$). Spatially, SCAT recovers occluded $F^O$ and injects $F^H$ via contact-aware cues. Fusing refined $f^O$ with joint and global contexts yields an aggregated motion token ($f^\text{token}$). Temporally, TIAT processes this token sequence to model temporal dynamics ($\hat{f}^\text{token}$), which MLPs decode into 4D HOI reconstruction.
  }
  \label{fig:framework}
\end{figure}

In this section, we first detail the preparation of human motion priors and interaction-centric image patches (Sec.~\ref{sec:preparation}), followed by the 3D feature encoding that generates unified geometry-aware representations for human and object vertices (Sec.~\ref{sec:encoder}). We then introduce the Spatial Contact-Aware Transformer (SCAT) (Sec.~\ref{sec:spatial}), which recovers occluded object representations and injects human interaction priors via contact-aware cues. Finally, we describe the Temporal Interact-Aware Transformer (TIAT) (Sec.~\ref{sec:temporal}), which ensures temporal consistency by modeling continuous interaction dynamics. An overview of our framework is illustrated in Fig.~\ref{fig:framework}.

\subsection{Data Preparation}
\label{sec:preparation}
We employ GVHMR~\cite{shen2024gvhmr} to estimate human motion from the input video $\mathcal{I}$. We select GVHMR for its capability to recover robust global human trajectories and temporally coherent motion, which provides a stable initialization for subsequent interaction modeling. Notably, as our task operates under a fixed-camera assumption, we disable its camera tracking module (DPVO~\cite{teed2024deep,lipson2024deep}) and establish the camera coordinate system directly as our global world reference frame. Specifically, given the input video, GVHMR estimates the intrinsic camera parameters $K \in \mathbb{R}^{3\times 3}$ and predicts both initial world-grounded human vertices $V^H_t \in \mathbb{R}^{N_h \times 3}$ and human joints $V^J_t \in \mathbb{R}^{N_j \times 3}$ for each frame $t$.

\paragraph{Interaction-Centric Crop.} To effectively decouple the human’s global translation from the local HOI reconstruction task, we generate interaction-centric image patches. Specifically, for each frame $t$, we compute a square bounding box $\mathbf{B}_t$ centered at $p_{pelvis}$, which is the 2D projection of the 3D pelvis joint $V_{pelvis}^{J}$, dynamically scaling it based on the union of input human and object masks to ensure the entire interaction context is fully visible. Subsequently, we crop the original frames based on their respective $\mathbf{B}_t$ and resize them to obtain a fixed-resolution patch sequence $\mathcal{I}^{crop} = \{I_t^{crop} \in \mathbb{R}^{S \times S \times 5}\}_{t=1}^N$. Concurrently, we dynamically update the original camera intrinsics $K$ to the cropped intrinsics $K^{crop}_t$ by applying the 2D translation offset of the bounding box and scaling the focal length and principal point to match the target resolution $S \times S$.

Finally, the pre-processed data tuple $\{\mathcal{I}^{crop}, \mathbf{B}, V^H, V^J, K^{crop}\}$ serves as the input for the subsequent encoding stage.

\subsection{3D Vertex Encoder}\label{sec:encoder}
The 3D Vertex Encoder is designed to extract unified, geometry-aware feature representations for human joints, human vertices, and initialized object vertices from each cropped video frame.
\paragraph{Visual Feature Extraction and Object Initialization.}First, we feed the RGB components of the interaction-centric image patch $I^{crop}_t$ into a frozen, pre-trained ViT backbone~\cite{dosovitskiy2020vit} (DINOv3~\cite{simeoni25dinov3} ViT-L/16), yielding a dense visual feature map $\mathcal{F}^{img}_t \in \mathbb{R}^{h \times w \times C}$. To initialize the 3D geometry of the object at frame $t$, we perform masked average pooling on $\mathcal{F}^{img}_t$ utilizing the corresponding object mask to aggregate global object cues. This pooled feature is then passed through an MLP decoder to regress the initial object rotation $R_t \in SO(3)$ and translation $T_t \in \mathbb{R}^3$ relative to the human coordinate system. By applying these rigid body transformations alongside the human-to-world transformation to the vertices of the canonical object template $V^{temp}$, we obtain the initial object vertices $V^O_t \in \mathbb{R}^{N_o \times 3}$ for subsequent feature projection.
\paragraph{Unified Projection-based Feature Sampling.} With the initialized object vertices $V^O_t$ and the human geometry ($V^H_t, V^J_t$) obtained in Sec.~\ref{sec:preparation}, we adopt a unified projection-based sampling strategy to encode all entities. Specifically, given the camera intrinsics $K^{crop}_t$, Assuming identity extrinsics since the camera acts as the world origin, we project each 3D coordinate $v_t \in V^H_t \cup V^J_t \cup V^O_t$ onto the 2D image plane:
$$p_t = \Pi(K^{crop}_t, v_t),$$
where $\Pi$ denotes the perspective projection. We then sample the visual features from $\mathcal{F}^{img}_t$ at location $p_t$ via bilinear interpolation. To preserve explicit 3D spatial awareness, we concatenate the sampled visual features with the vertex's corresponding 3D coordinates $(x, y, z)$. This process yields a fused feature vector $F_t \in \mathbb{R}^{C+3}$ for every point. Consequently, for each frame $t$, we obtain robust feature sets for human vertices $F^H_t \in \mathbb{R}^{N_h \times (C+3)}$, human joints $F^J_t \in \mathbb{R}^{N_j \times (C+3)}$, and object vertices $F^O_t \in \mathbb{R}^{N_o \times (C+3)}$. These point-level representations, which encapsulate both semantic context and spatial geometry, serve as the input for the subsequent interaction learning modules.

\subsection{Spatial Contact-Aware Transformer}
\label{sec:spatial}
To mitigate object feature degradation caused by hand occlusions, we introduce the Spatial Contact-Aware Transformer (SCAT). This module leverages the robust human mesh prior to refine object representations through two sequential stages: occlusion-aware internal refinement and human-guided contact injection.
\paragraph{Occlusion-Aware Internal Refinement.}First, we apply Multi-Head Self-Attention (MHSA)~\cite{vaswani2017attention} to the initial object features $F^O_t$ to reconstruct occluded regions by aggregating global context from the visible vertices. This internal refinement ensures feature consistency across the entire object surface. A standard Feed-Forward Network (FFN) subsequently processes the attended features:
$$\tilde{F}^O_t = \text{LN}(F^O_t + \text{MHSA}(F^O_t, F^O_t, F^O_t)), \quad \hat{F}^O_t = \text{LN}(\tilde{F}^O_t + \text{FFN}(\tilde{F}^O_t)),$$
where $\text{LN}$ denotes Layer Normalization.
\paragraph{Human-Guided Contact Injection.}Next, we inject human interaction priors via a single-layer Multi-Head Cross-Attention (MHCA) mechanism. Treating the internally refined object features $\hat{F}^O_t$ as queries, and the human features $F^H_t$ as keys and values, we compute the enhanced output $f^O_t$:
$$A = \text{Softmax}\left(\frac{(\hat{F}^O_t W^q)(F^H_t W^k)^\top}{\sqrt{d_k}}\right),$$
$$ \tilde{f}^O_t = \text{LN}(\hat{F}^O_t + A (F^H_t W^v)), \quad f^O_t = \text{LN}(\tilde{f}^O_t + \text{FFN}(\tilde{f}^O_t)).$$
Crucially, the cross-attention matrix $A$ implicitly captures a dual consistency: it highlights human vertices that are both \textit{semantically corresponding} (e.g., hand-to-handle) and \textit{spatially proximal}. This efficient single-layer architecture allows the object vertices to "borrow" clean, unoccluded features from the interacting body parts, thereby correcting occlusion-induced noise without introducing heavy computational overhead.

\subsection{Temporal Interact-Aware Transformer}
\label{sec:temporal}
While the spatial module resolves per-frame occlusion based on geometric proximity, independent predictions often exhibit high-frequency jitter and lack physical coherence. To ensure temporal consistency, this module explicitly models the kinematic correlations between human joints and the manipulated object. By attending to cross-frame interaction dynamics, it leverages the temporal context of the entire $N$-frame sequence to correct and refine current-frame object representations. Concurrently, the integration of global spatial context grounds these local interactions into smooth, absolute world trajectories.
\paragraph{Global Motion Tokenization.} We first condense the spatially refined features into a compact frame-level representation. To explicitly perceive skeletal kinematics and recover absolute trajectories decoupled by our interaction-centric cropping, we project the human joint features $F^J_t$ via a linear layer to obtain $f^J_t$. We then apply average pooling across the vertex dimensions for both $f^J_t$ and the refined object features $f^O_t$. Finally, we fuse these complementary cues—local interaction context, skeletal motion, and a global context embedding $f^g_t$ (derived by projecting the concatenation of the bounding box $\mathbf{B}_t$ and the 3D pelvis coordinates $V^J_{\text{pelvis}, t}$ via a linear layer)—into a unified latent space via separate MLPs and element-wise summation:$$f^{\text{token}}_t = \text{MLP}_O(\text{AvgPool}(f^O_t)) + \text{MLP}_J(\text{AvgPool}(f^J_t)) + \text{MLP}_G(f^g_t),$$
where $f^{\text{token}}_t \in \mathbb{R}^{512}$ serves as a robust, interact-aware motion descriptor for the subsequent temporal aggregation.
\paragraph{Temporal Interact-Aware Transformer.}The token sequence is processed by a multi-layer Transformer encoder~\cite{vaswani2017attention} to capture temporal motion dependencies. While GVHMR~\cite{shen2024gvhmr} employs similar temporal mechanisms for pure human motion recovery, our framework adapts these strategies specifically to process unified HOI tokens, thereby enforcing physical coherence between interacting entities. Formally, let $f^{(l-1)}_t$ denote the input to the $l$-th transformer layer, initialized as $f^{(0)}_t = f^{\text{token}}_t$. We integrate Rotary Positional Embeddings (RoPE)~\cite{su2021roformer} and a Local Attention Mask into a unified self-attention formulation:
$$\tilde{f}^{(l)}_t = \sum_{s=1}^N \text{Softmax}_s \Big( (\mathbf{R}^p_t \mathbf{W}^q f^{(l-1)}_t)^\top (\mathbf{R}^p_s \mathbf{W}^k f^{(l-1)}_s) + \mathbf{m}_{ts} \Big) \mathbf{W}^v f^{(l-1)}_s,$$
where $\mathbf{W}^q, \mathbf{W}^k, \mathbf{W}^v$ denote learnable projection matrices, and $\mathbf{R}^p_t, \mathbf{R}^p_s$ are rotary embedding operators at positions $t$ and $s$, preserving translation invariance through relative distance encoding. For robust generalization to arbitrarily long sequences, the mask $\mathbf{m}_{ts}$ limits the receptive field to a training window $L$: $\mathbf{m}_{ts} = 0$ if $|t - s| < L$, and $-\infty$ otherwise. The attended feature $\tilde{f}^{(l)}_t$ is subsequently processed by standard Layer Normalization and Feed-Forward Networks (FFN) to yield the layer output $f^{(l)}_t$. The final temporally coherent feature is extracted from the last layer $L_{temp}$, \textit{i.e.}, $\hat{f}^{\text{token}}_t = f^{(L_{temp})}_t$.
\paragraph{Parameter Regression.} Finally, the temporally coherent features $\{\hat{f}^{\text{token}}_t\}_{t=1}^N$ are decoded by lightweight heads to produce the 4D HOI reconstruction directly in the world coordinate system. The human branch regresses the SMPL~\cite{loper2023smpl,romero2017mano} parameters (global orientation, body pose $\hat{\theta}$, and shape $\hat{\beta}$) to recover the world-grounded mesh sequence $\{\hat{M}^h_t\}_{t=1}^N$. Simultaneously, the object branch predicts the absolute 6-DoF pose trajectory—parameterized by global rotation $\{\hat{R}_t\}_{t=1}^N \in SO(3)^N$ and global translation $\{\hat{T}_t\}_{t=1}^N \in \mathbb{R}^{3\times N}$—to align the canonical object template $V_{temp}$ in the world space. By effectively integrating spatial contact priors with temporal contexts, our method directly yields accurate and stable 4D sequences without requiring expensive test-time optimization.
\subsection{Loss functions}
\label{sec:loss}
We train THO end-to-end by minimizing the overall objective $\mathcal{L}$:
\begin{equation}
\mathcal{L} = \mathcal{L}_\text{param} + \mathcal{L}_\text{mesh} + \mathcal{L}_\text{acc}
\end{equation}
Here, all terms compute the $L_1$ distance against their respective ground truth. Specifically, $\mathcal{L}_\text{param}$ is calculated on the predicted SMPL~\cite{loper2023smpl,romero2017mano} parameters ($\hat{\theta}$, $\hat{\beta}$) and global 6-DoF object pose ($\hat{R}$, $\hat{T}$). $\mathcal{L}_\text{mesh}$ is applied to the vertex coordinates of the human $\hat{M}^h$ and object $\hat{M}^o$, including an edge length consistency term for $\hat{M}^h$~\cite{nam2024contho}. Finally, $\mathcal{L}_\text{acc}$ measures the velocity and acceleration of all 3D vertices. Please refer to the supplementary material for details.

%% file: paragraphs/4_Experiments.tex
\section{Experiments}
\label{sec:Experiments}

\subsection{Benchmark}
\label{sec:benchmark}

\noindent\textbf{Datasets.} We evaluate THO on two public datasets: (1) BEHAVE Dataset~\cite{bhatnagar2022behave}, which captures 7 subjects interacting with 20 different objects in indoor environments. We utilize the extended 30 fps annotations and follow VisTracker~\cite{xie2023vistracker} for the data split to ensure fair comparison. (2) InterCap Dataset~\cite{huang2022intercap}, which contains interactions between 10 subjects and 10 diverse objects with registrations at 30 fps. Similarly, we adopt the exact same split setting as VisTracker.

\noindent\textbf{Evaluation metrics.} Following previous works~\cite{xie2023vistracker,xie2025cari4d}, we evaluate the reconstruction quality of the human, object, and combined meshes using Chamfer distance, denoted as $\text{CD}_{h}$, $\text{CD}_{o}$, and $\text{CD}_{c}$, respectively. To further assess the geometric fidelity of symmetric objects, we define the Vertex-to-Vertex ($\text{V2V}$) metric as the mean distance error of object vertices against the ground truth. To transform the predictions into a unified coordinate system and rigorously evaluate the temporal consistency of the reconstructions, we align the combined human and object mesh of the first frame to the ground truth and apply this exact transformation to the entire video sequence~\cite{xie2025cari4d}. Both the $\text{CD}$ and $\text{V2V}$ metrics are then computed based on this globally aligned sequence. Finally, to evaluate the temporal smoothness of the reconstructed sequences, we define $\text{Acc}_{h}$ and $\text{Acc}_{o}$ as the mean acceleration error for the human and object vertices against the ground truth, respectively.

\subsection{Implementation Details}
\label{sec:implementation}

We implement our THO framework using PyTorch~\cite{paszke2019pytorch}. The model is optimized via AdamW~\cite{loshchilov2017decoupled} with a weight decay of $1 \times 10^{-2}$ and an initial learning rate of $1 \times 10^{-4}$ that gradually decays to $1 \times 10^{-6}$. For the backbone, we utilize GVHMR~\cite{shen2024gvhmr} excluding its DPVO~\cite{teed2024deep,lipson2024deep} module and a frozen pre-trained DINOv3-ViT-L/16~\cite{simeoni25dinov3}. Notably, we zero-initialize $\text{MLP}_G$ to facilitate TIAT's progressive learning, while the remaining modules are randomly initialized. The training spans 150 epochs with a batch size of 16 and a mask sequence length of $L=64$, taking approximately 27 hours on 8 RTX 4090 GPUs.

\begin{figure}[t]
    \centering
    \includegraphics[width=\linewidth]{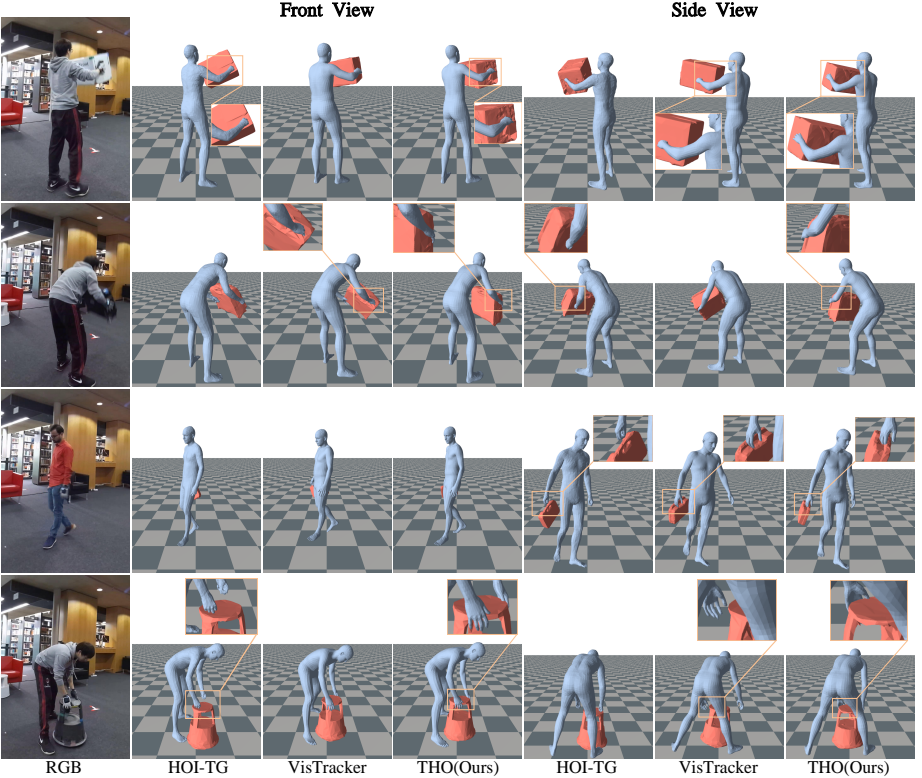}
    \caption{\textbf{Qualitative comparison on BEHAVE~\cite{bhatnagar2022behave}.} THO mitigates inaccurate symmetric object pose predictions (Rows 1-2), yields more reasonable poses under severe occlusion (Row 3), and ensures better human-object contact (Row 4).}
    \label{fig:qualitative}
\end{figure}

\subsection{Comparison with State-of-the-art Methods}

\begin{table}[tb]
  \caption{\textbf{Quantitative comparison with state-of-the-art methods on BEHAVE~\cite{bhatnagar2022behave} and InterCap~\cite{huang2022intercap} datasets.} Reconstruction accuracy metrics ($\text{CD}$ and $\text{V2V}$) are scaled in cm, temporal smoothness  metrics ($\text{Acc}$) in cm/s$^2$, and inference speed in FPS. The best results are highlighted in bold.}
  \centering
  \resizebox{\textwidth}{!}{
  \begin{tabular}{@{}l|cccccc|cccccc|c@{}}
    \toprule
    \raisebox{-2.2ex}[0pt][0pt]{\bf Method} & \multicolumn{6}{c|}{\bf BEHAVE} & \multicolumn{6}{c|}{\bf InterCap} & \raisebox{-2.2ex}[0pt][0pt]{\bf FPS$\uparrow$} \\ 
    \cmidrule{2-7} \cmidrule{8-13}
     & $\text{CD}_{h}\downarrow$ & $\text{CD}_{o}\downarrow$ & $\text{CD}_{c}\downarrow$ & $\text{V2V}\downarrow$ & $\text{Acc}_{h}\downarrow$ & $\text{Acc}_{o}\downarrow$ & $\text{CD}_{h}\downarrow$ & $\text{CD}_{o}\downarrow$ & $\text{CD}_{c}\downarrow$ & $\text{V2V}\downarrow$ & $\text{Acc}_{h}\downarrow$ & $\text{Acc}_{o}\downarrow$ & \\
    \midrule
    
    CHORE~\cite{xie2022chore}       & 25.82 & 28.94 & 26.52 & 38.16 & 2.75 & 4.59 & 25.14 & 32.45 & 27.31 & 41.72 & 2.61 & 4.37 & 0.07 \\
    HOI-TG~\cite{wang2025endtoendhoi}      & 21.19 & 24.70 & 22.67 & 34.69 & 2.44 & 3.76 & 23.52 & 26.63 & 24.20 & 36.94 & 2.47 & 3.64 & 4.81 \\
    VisTracker~\cite{xie2023vistracker}  & 11.24  & \textbf{12.58}  & 11.63  & 22.45 & 0.83 & 1.21 & 12.57  & \textbf{13.80} & 12.76  & 24.35 & 0.86 & 1.27 & 0.05 \\
    Ours        & \textbf{8.49}  & 13.36  & \textbf{10.06}  & \textbf{21.68} & \textbf{0.79} & \textbf{1.02} & \textbf{9.31}  & 14.45 & \textbf{11.51}  & \textbf{23.72} & \textbf{0.83} & \textbf{1.12} & \textbf{31.5} \\
    \bottomrule
  \end{tabular}
  }
  \label{tab:sota_comparison}
\end{table}

\begin{figure}[tb]
  \centering
  \includegraphics[height=4.5cm]{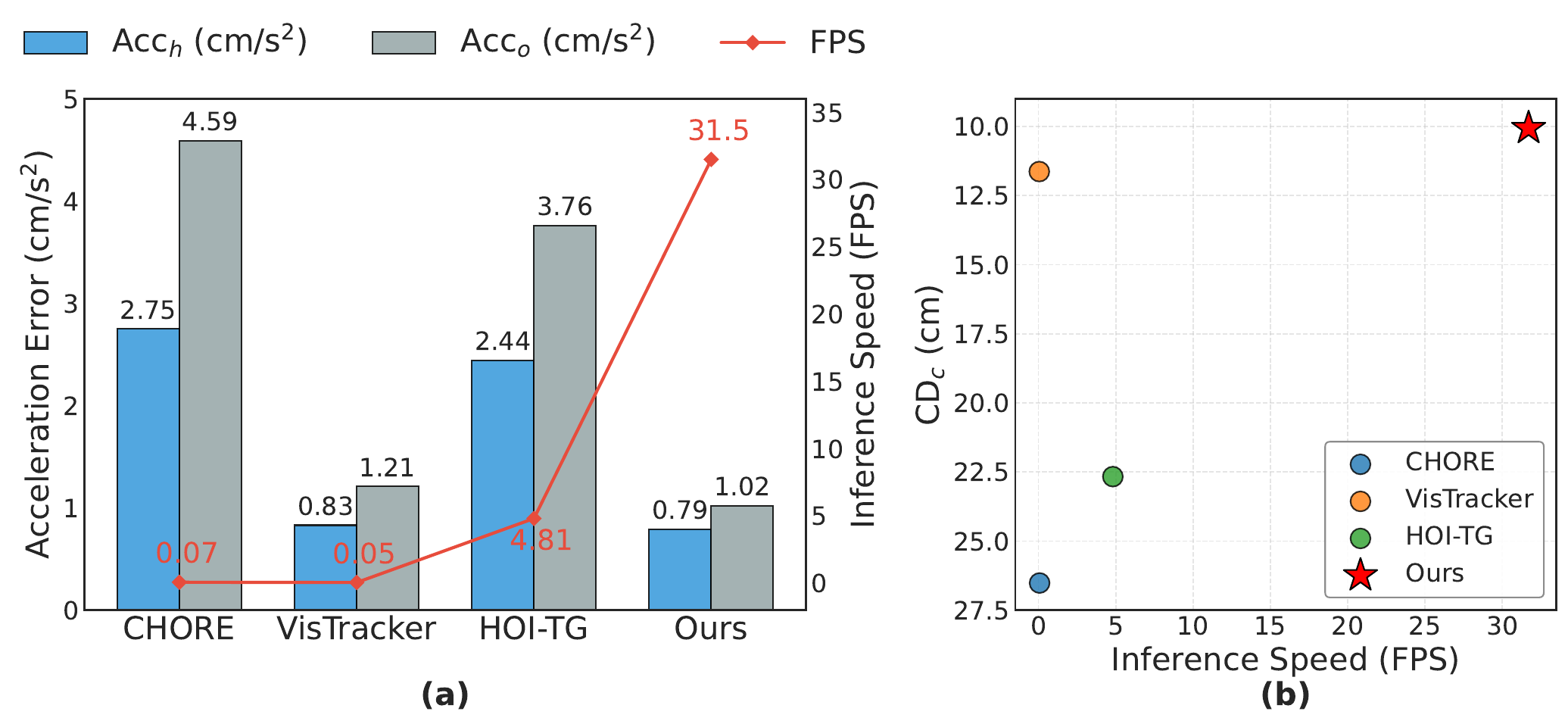}
    \caption{\textbf{Efficiency vs. Performance on BEHAVE.} THO achieves real-time inference (31.5 FPS), outperforming all baselines not only in inference speed but also yielding the best temporal smoothness ($\text{Acc}_h, \text{Acc}_o$) as shown in \textbf{(a)}, and the highest reconstruction accuracy ($\text{CD}_c$) as shown in \textbf{(b)}.}
  \label{fig:efficiency}
\end{figure}

\noindent\textbf{Reconstruction Accuracy.} As shown in Table~\ref{tab:sota_comparison}, THO achieves state-of-the-art reconstruction accuracy across both datasets, outperforming recent baselines (CHORE~\cite{xie2022chore}, HOI-TG~\cite{wang2025endtoendhoi}, and VisTracker~\cite{xie2023vistracker}). Specifically, THO improves the Chamfer Distance of combined mesh ($\text{CD}_{c}$) by 13\% and Vertex-to-Vertex ($\text{V2V}$) metrics by 3.4\% on the BEHAVE dataset~\cite{bhatnagar2022behave}. Qualitatively, THO robustly addresses two challenging scenarios that typically degrade baseline performance. As demonstrated in the top half of Fig.~\ref{fig:qualitative}, symmetric objects inherently introduce multi-solution ambiguities. Consequently, single-frame 3D methods lack temporal context and frequently yield inconsistent predictions (e.g., flipped object poses), whereas the 4D optimization in VisTracker heavily relies on accurate pose initialization. In contrast, our Temporal Interact-Aware Transformer (TIAT) captures continuous relative dynamics to enforce long-term motion coherence, effectively resolving these per-frame geometric ambiguities. Furthermore, as demonstrated in the bottom half of Fig.~\ref{fig:qualitative}, under severe hand occlusions, our Spatial Contact-Aware Transformer (SCAT) injects human interaction priors based on semantic and spatial proximity. This explicitly corrects occlusion noise, recovering physically realistic geometry while ensuring higher-quality human-object contact where baseline methods fail.

\noindent\textbf{Temporal Smoothness.} THO achieves the lowest acceleration errors ($\text{Acc}_h$, $\text{Acc}_o$) across both benchmarks. Single-frame baselines (CHORE~\cite{xie2022chore} and HOI-TG~\cite{wang2025endtoendhoi}) inherently lack temporal coherence due to independent per-frame processing. Our method overcomes this limitation by explicitly modeling continuous motion dynamics through the Temporal Interact-Aware Transformer (TIAT). Furthermore, while video-based methods like VisTracker~\cite{xie2023vistracker} employ segment-based optimization for temporal smoothness, their inference strategy on long videos inevitably introduces subtle motion discontinuities at segment boundaries. In contrast, THO ensures seamless temporal transitions across arbitrarily long sequences via a single feed-forward pass equipped with a local attention mask, setting a new standard for stable 4D HOI reconstruction.

\noindent\textbf{Inference Efficiency.} A critical bottleneck of prior 4D HOI pipelines is expensive test-time optimization. As shown in Fig.~\ref{fig:efficiency}, evaluated on a single RTX 4090 GPU, VisTracker~\cite{xie2023vistracker} (0.05 FPS) and CHORE~\cite{xie2022chore} (0.07 FPS) operate far below interactive rates. Furthermore, HOI-TG~\cite{wang2025endtoendhoi}, the fastest feed-forward 3D baseline, is constrained to 4.81 FPS. In contrast, our end-to-end architecture achieves \textbf{31.5 FPS}, delivering over $6.5 \times$ speedup over the fastest 3D method and $>600\times$ acceleration over VisTracker. By establishing a new accuracy-latency paradigm, THO pioneers real-time 4D HOI reconstruction, unlocking interactive applications previously unattainable by high-latency pipelines.

\subsection{Ablation Study}
\label{sec:ablation}

To validate our core components, we conduct comprehensive ablation studies on the BEHAVE dataset~\cite{bhatnagar2022behave}. We investigate the contributions of the Spatial Contact-Aware Transformer (SCAT), the Temporal Interact-Aware Transformer (TIAT), and the Global-Local Motion Disentanglement (GLMD) strategy.

\begin{table}[tb]
  \caption{\textbf{Ablation study of SCAT.} We evaluate the progressive contributions of Occlusion-Aware Internal Refinement (OR) and Human-Guided Contact Injection (HI) to refining object geometry and enhancing contact precision.}
  \centering
  \setlength{\tabcolsep}{2pt}
  \renewcommand{\arraystretch}{0.8}
  \small 
  \resizebox{0.50\textwidth}{!}{
  \begin{tabular}{@{}lcccc@{}}
    \toprule
    & $\text{CD}_{h} \downarrow$ & $\text{CD}_{o} \downarrow$ & $\text{CD}_{c} \downarrow$ & $\text{V2V} \downarrow$ \\
    \midrule
    w/o SCAT & 8.53 & 14.59 & 10.47 & 24.32 \\
    w/ OR & 8.52 & 14.22 & 10.35 & 23.08 \\
    w/ OR\&HI & \textbf{8.49} & \textbf{13.36} & \textbf{10.06} & \textbf{21.68} \\
    \bottomrule
  \end{tabular}
  }
  \label{tab:ablation_SCAT}
\end{table}

\begin{figure}[t]
    \centering
    \includegraphics[width=0.77\linewidth]{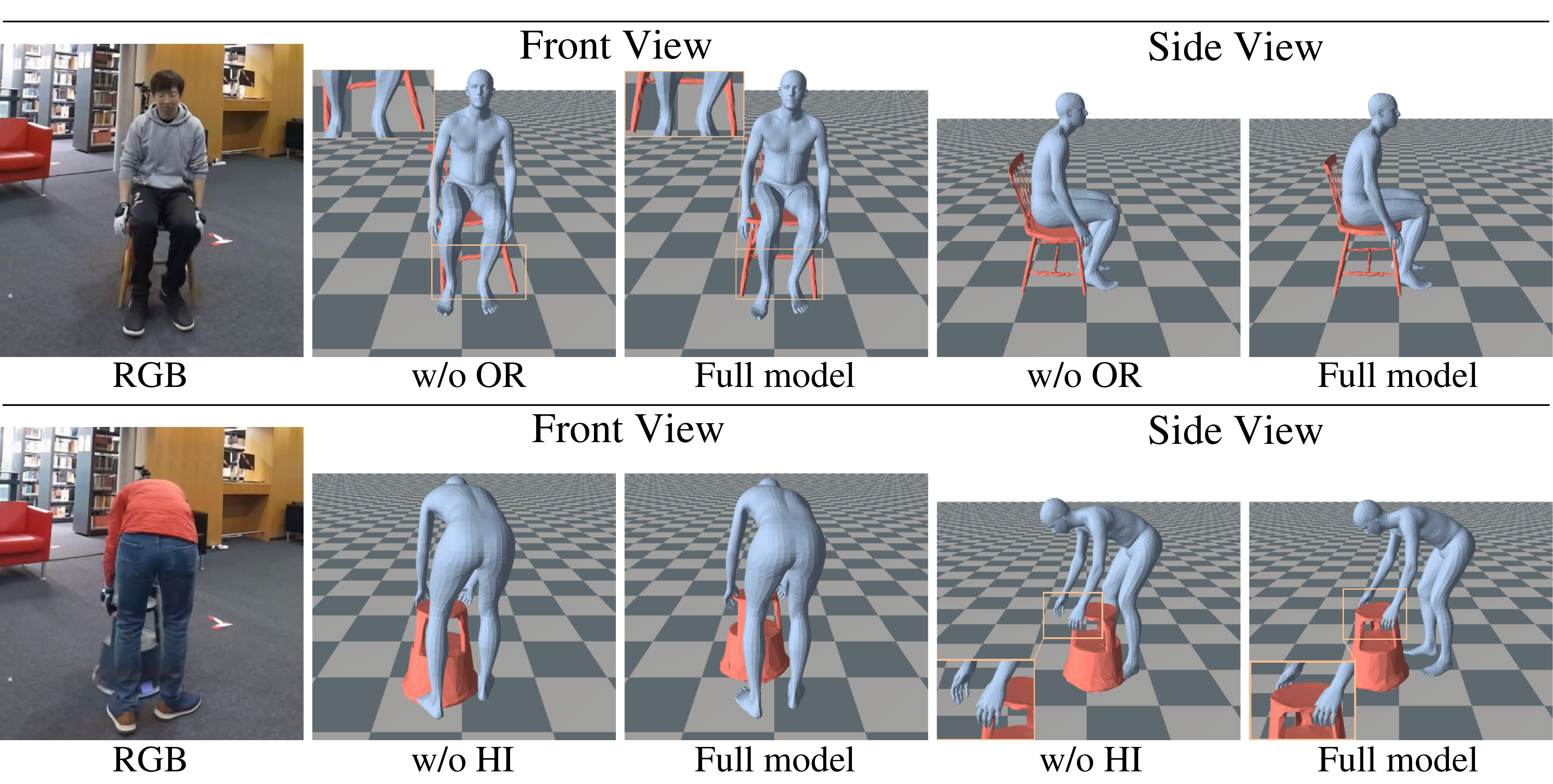}
    \caption{\textbf{Impact of SCAT.} OR recovers complete object geometry from occluded views, while HI corrects the relative pose to ensure physically consistent human-object contact, eliminating the misalignment observed in the baseline.}
    \label{fig:ablation_occlusion}
\end{figure}

\begin{figure}[t]
    \centering
    \includegraphics[width=0.82\linewidth]{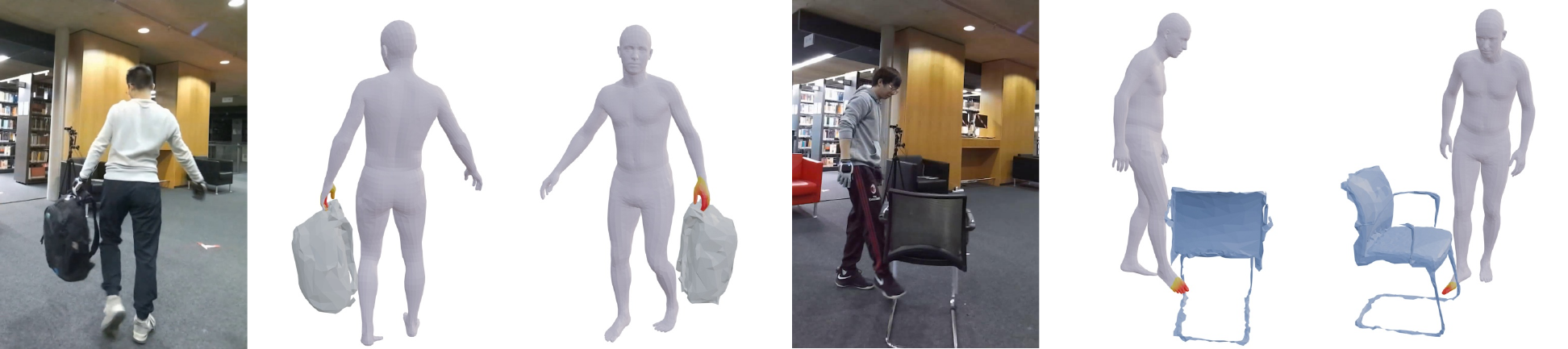}
    \caption{\textbf{Visualization of Human-Guided Contact Injection.} Left: Attention highlights the interacting hand while suppressing the proximal calf (semantic mismatch). Right: It isolates the contacting foot, ignoring the distant non-interacting foot.}
    \label{fig:ablation_heatmap}
\end{figure}

\noindent\textbf{Impact of Spatial Contact-Aware Transformer.}
We investigate the impact of the two sub-modules within the Spatial Contact-Aware Transformer (SCAT), as detailed in Table~\ref{tab:ablation_SCAT}. The first row presents the baseline where the SCAT is entirely removed (i.e., $f^O = F^O$). The second row corresponds to the addition of Occlusion-Aware Internal Refinement (OR), while the third row shows further integration of Human-Guided Contact Injection (HI). Improvements in object-related metrics ($\text{CD}_{o}$ and $\text{V2V}$) demonstrate that both modules effectively enhance object motion reconstruction. As illustrated in Fig.~\ref{fig:ablation_occlusion}, compared to baseline, OR improves the quality of reconstruction under partial occlusion. Furthermore, HI enables object vertices to perceive spatially and semantically proximal human vertices, resulting in more plausible contact at interaction.

Similar to recent interaction-centric designs~\cite{nam2024contho, wang2025endtoendhoi}, to illustrate the learned contact priors, Fig.~\ref{fig:ablation_heatmap} visualizes the interaction heatmap derived by averaging the cross-attention weights assigned by all object queries to each human vertex. The visualization confirms that the model successfully focuses on semantically corresponding and spatially proximal human parts (e.g., the interacting hand), while avoiding regions that are either spatially close but semantically irrelevant, or semantically corresponding but distant, thereby validating that SCAT effectively captures the intended dual consistency of semantic and spatial proximity.

\begin{table}[tb]
  \caption{\textbf{Ablation study of TIAT and GLMD.} We assess the impact of removing full modules and their specific sub-components (RoPE in TIAT, $f^g$ in GLMD) on reconstruction accuracy and temporal coherence.
  }
  \centering
  \setlength{\tabcolsep}{2pt}
  \renewcommand{\arraystretch}{0.85}
  \small 
  \resizebox{0.65\textwidth}{!}{
  \begin{tabular}{@{}lcccccc@{}}
    \toprule
     & $\text{CD}_{h} \downarrow$ & $\text{CD}_{o} \downarrow$ & $\text{CD}_{c} \downarrow$ & $\text{V2V} \downarrow$ & $\text{Acc}_{h} \downarrow$ & $\text{Acc}_{o} \downarrow$ \\
    \midrule
    w/o TIAT & 10.34 & 16.45 & 12.38 & 27.31 & 1.12 & 1.97 \\
    w/o RoPE & 8.65 & 13.60 & 10.27 & 22.74 & 0.84 & 1.20 \\
    \midrule
    w/o GLMD & 8.63 & 15.26 & 10.79 & 25.85 & 0.81 & 1.16 \\
    w/o $f^g$ & 14.70 & 21.32 & 17.12 & 31.55 & 1.67 & 2.75 \\
    \midrule
    full model & \textbf{8.49} & \textbf{13.36} & \textbf{10.06} & \textbf{21.68} & \textbf{0.79} & \textbf{1.02} \\
    \bottomrule
  \end{tabular}
  }
  \label{tab:ablation_all}
\end{table}

\noindent\textbf{Impact of Temporal Interact-Aware Transformer.}
We evaluate the Temporal Interact-Aware Transformer (TIAT) in the top rows of Table~\ref{tab:ablation_all}. Removing TIAT (i.e., $\hat{f}^{token} = f^{token}$) degrades the model to single-frame prediction. Despite the robust absolute spatial priors provided by the human motion baseline, the reconstruction quality notably declines. Crucially, the lack of explicit temporal modeling leads to a marked increase in acceleration error, manifesting as high-frequency jitter. This indicates that spatial priors alone are insufficient for coherent 4D reconstruction, whereas TIAT suppresses jitter by capturing continuous interaction dynamics. To investigate the efficacy of different positional encoding and inference schemes, we also evaluated a variant using Absolute Positional Embeddings (APE)~\cite{vaswani2017attention} combined with a sliding window inference strategy~\cite{kocabas2020vibe, choi2021tcmr}. The comparison confirms that RoPE~\cite{su2021roformer} better captures intrinsic relative motion dynamics, whereas our inference strategy with Local Attention Mask ensures robust generalization to arbitrary sequence lengths.

\noindent\textbf{Impact of Global-Local Motion Disentanglement.}
Finally, we evaluate the strategy of decoupling local interaction from global trajectory (GLMD) in the middle rows of Table~\ref{tab:ablation_all}.
First, regarding the Interaction-Centric Crop (\cref{sec:preparation}), the baseline w/o GLMD operates on full input frames instead of generating interaction-centric image patches. The degradation in geometric metrics ($\text{CD}_{o}$ and $\text{V2V}$) confirms that implicitly learning high-frequency interaction details from global views is inefficient, motivating our cropping strategy to focus on fine-grained local dynamics. Second, we verify the necessity of Global Motion Tokenization (\cref{sec:temporal}) in recovering the spatial context lost during cropping. As shown in the row w/o $f^g$, omitting the global context embedding (which explicitly encodes the bounding box and pelvis coordinates) deprives the network of absolute trajectory information. Consequently, the model fails to anchor the interaction sequence in the world coordinate system, resulting in a significant increase in reconstruction errors across all spatially aligned metrics.

\noindent\textbf{Inference Speed of Individual Modules.} We analyze the inference time on a 1430-frame sequence using a single RTX 4090 GPU. The majority of the computational overhead stems from the human motion estimator and visual backbone. Specifically, GVHMR\cite{shen2024gvhmr} (without DPVO~\cite{teed2024deep,lipson2024deep}) takes 35.0 seconds, and DINOv3~\cite{simeoni25dinov3} feature extraction takes 8.4 seconds. Conversely, the remaining components of THO process the sequence in just 1.98 seconds. As a result, the complete pipeline runs at roughly 31.5 FPS, demonstrating that our proposed method ensures real-time performance with minimal added cost.

%% file: paragraphs/5_Conclusion.tex
\section{Discussions}
\label{sec:Conclusion}

\textbf{Conclusion.} In this paper, we presented THO, a novel end-to-end framework for real-time 4D HOI reconstruction. By devising a Spatial Contact-Aware Transformer and a Temporal Interact-Aware Transformer, our approach explicitly integrates human-guided contact priors and continuous interaction dynamics. This dual focus effectively mitigates the pose ambiguities caused by severe occlusions and significantly enhances the temporal coherence of the reconstructed motion. Extensive evaluations on the BEHAVE and InterCap benchmarks demonstrate that THO surpasses state-of-the-art methods in both reconstruction accuracy and motion smoothness. Crucially, with an inference speed of 31.5 FPS, our feed-forward pipeline achieves a substantial speedup over traditional optimization-centric methods, paving the way for real-time interactive applications. 

\noindent\textbf{Limitations and future work.} Our method currently exhibits a limitation when handling elongated objects. For such geometries, minor 3D pose variations often induce drastic changes in their 2D perspective projections, thereby disrupting the visual feature consistency essential for stable feed-forward prediction. In future work, we aim to overcome this vulnerability by exploring rotation-equivariant visual representations to further stabilize the reconstruction process.

%% file: paragraphs/6_details.tex
\section{More details of our THO}
\label{sec:details}

\subsection{Network architecture}
\label{sec:network}

\paragraph{3D Vertex Encoder.} 
The encoder independently processes the point-level features for the object, human, and skeleton joints. Specifically, the $1024$-D visual features sampled from the feature maps are concatenated with their $3$D coordinates. This $1027$-D geometry-aware representation is then projected into the shared $512$-D latent space via separate Multi-Layer Perceptrons (MLPs), ensuring that critical spatial relationships are preserved alongside semantic contexts.

\paragraph{Spatial Contact-Aware Transformer (SCAT).}
As detailed in the main text, SCAT refines the object representation through two sequential stages: Occlusion-Aware Internal Refinement (via Multi-Head Self-Attention~\cite{vaswani2017attention}) and Human-Guided Contact Injection (via Spatial Cross-Attention~\cite{vaswani2017attention}). Structurally, both attention mechanisms share identical hyperparameters, each utilizing $8$ attention heads within a $512$-D latent space. These sequential operations are subsequently processed by a unified Feed-Forward Network (FFN) with an expansion ratio of $4.0$ (yielding a hidden dimension of $2048$). This straightforward architecture effectively enables the network to integrate global object contexts and human interaction priors without introducing heavy computational overhead.

\paragraph{Global Motion Tokenization.}
Before temporal aggregation, the refined object features, the 3D human joints, and the global spatial context are fused into a unified token space. As detailed in the main text, the global context is represented by a 6-D vector concatenating the 3D pelvis coordinates and the 3 parameters of the square bounding box (center $x$, center $y$, and side length $s$). These three complementary cues are encoded into a $512$-D token ($f_{t}^{token}$) via separate MLPs and element-wise summation. Notably, the final linear layer of the global context embedder ($MLP_{G}$) is strictly zero-initialized. This specific structural bias is designed to encourage the network to prioritize learning fine-grained local interactions during early training, gradually integrating the global absolute trajectory as the optimization advances.

\paragraph{Temporal Interact-Aware Transformer (TIAT).}
TIAT comprises $12$ Transformer encoder layers. Each layer features $8$ attention heads and an FFN hidden dimension of $2048$. To effectively capture continuous interaction dynamics, we employ Rotary Positional Embeddings (RoPE)~\cite{su2021roformer} to encode relative temporal distances. Furthermore, a sliding local attention mask with a window size of $L=64$ is applied during training. This enforces temporal translation invariance and enables robust generalization to arbitrarily long video sequences during inference.

\subsection{Loss functions}
\label{sec:loss} 

In the main paper, the overall training objective is formulated as $\mathcal{L} = \mathcal{L}_{param} + \mathcal{L}_{mesh} + \mathcal{L}_{acc}$. We explicitly decompose these terms into their respective sub-losses and detail their empirically determined weights below.

\paragraph{Parameter Loss ($\mathcal{L}_{param}$).}
This term supervises the parametric representations of the human and the object ($\hat{R}, \hat{T}$). Note that the human kinematic predictions comprehensively encapsulate the SMPL~\cite{loper2023smpl,romero2017mano} bodily rotations ($\hat{\theta}$, including global orientation and joint poses), shape ($\hat{\beta}$), as well as the global spatial translation. To ensure coherent spatial alignment, we explicitly decompose the human supervision into a rotation loss ($\mathcal{L}_{rot}^h$) and a translation loss ($\mathcal{L}_{trans}^h$). The overall parameter loss is formulated as:
$$ \mathcal{L}_{param} = \lambda_{rot}^h \mathcal{L}_{rot}^h + \lambda_{\beta}^h \mathcal{L}_{\beta}^h + \lambda_{trans}^h \mathcal{L}_{trans}^h + \lambda_{rot}^o \mathcal{L}_{rot}^o + \lambda_{trans}^o \mathcal{L}_{trans}^o $$
To avoid gimbal lock, all rotations (for both the human and the object) are mapped to the continuous 6D rotation space (R6D)~\cite{zhou2019continuity} and supervised via the $L_1$ norm ($\mathcal{L}_{rot}^h, \mathcal{L}_{rot}^o$). The human shape ($\mathcal{L}_{\beta}^h$) and the translations ($\mathcal{L}_{trans}^h, \mathcal{L}_{trans}^o$) are independently supervised via the $L_1$ norm. The weights are empirically set to balance the scale differences between rotation representations and metric translations: $\lambda_{rot}^h = 0.2$, $\lambda_{\beta}^h = 0.2$, $\lambda_{trans}^h = 1.0$, $\lambda_{rot}^o = 0.2$, and $\lambda_{trans}^o = 1.0$.

\paragraph{Mesh and Geometry Loss ($\mathcal{L}_{mesh}$).}
This term ensures fine-grained spatial alignment and anchors the human-object interaction. It is formulated as a weighted sum of five geometric constraints:
$$ \mathcal{L}_{mesh} = \lambda_{v}^h \mathcal{L}_{v}^h + \lambda_{v}^o \mathcal{L}_{v}^o + \lambda_{J} \mathcal{L}_{J} + \lambda_{edge} \mathcal{L}_{edge} + \lambda_{rel} \mathcal{L}_{rel\_dist} $$
$\mathcal{L}_{v}^h$, $\mathcal{L}_{v}^o$, and $\mathcal{L}_{J}$ are the $L_1$ distances for human vertices, object vertices, and 3D skeleton joints, respectively. Following~\cite{nam2024contho}, $\mathcal{L}_{edge}$ penalizes human mesh edge length differences to prevent local distortions. $\mathcal{L}_{rel\_dist}$ applies an $L_1$ constraint on the 3D relative distance vector between the human pelvis joint and the object's translation center to maintain interaction semantics under occlusion. In our experiments, all geometric constraints contribute equally: $\lambda_{v}^h = \lambda_{v}^o = \lambda_{J} = \lambda_{edge} = \lambda_{rel} = 1.0$.

\paragraph{Temporal Smoothness Loss ($\mathcal{L}_{acc}$).}
To ensure long-term physical coherence and smooth the overall vertex trajectories as noted in the main text, $\mathcal{L}_{acc} = \mathcal{L}_{acc}^{h} + \mathcal{L}_{acc}^{o}$ explicitly constrains the first-order (velocity, $v$) and second-order (acceleration, $a$) temporal derivatives directly in the 3D vertex space. For both the human ($e=h$) and the object ($e=o$), the loss combines supervised $L_1$ differences against ground-truth kinematics with an unsupervised acceleration magnitude regularizer:
$$\mathcal{L}_{acc}^{e} = \lambda_{vel}^e ||\hat{v}_t^e - v_t^e||_1 + \lambda_{a}^e ||\hat{a}_t^e - a_t^e||_1 + \lambda_{reg}^e ||\hat{a}_t^e||_1$$
We empirically set the human temporal weights to $\lambda_{vel}^h = 0.5$, $\lambda_{a}^h = 0.1$, and $\lambda_{reg}^h = 1.0$. The object uses identical supervised weights ($\lambda_{vel}^o = 0.5$, $\lambda_{a}^o = 0.1$) but a smaller regularizer $\lambda_{reg}^o = 0.5$.

%% file: paragraphs/7_limitations.tex
\section{Extended Analysis of Limitations}
\label{sec:limitations}

\begin{figure}[tb]
  \centering
  \includegraphics[width=0.85\textwidth]{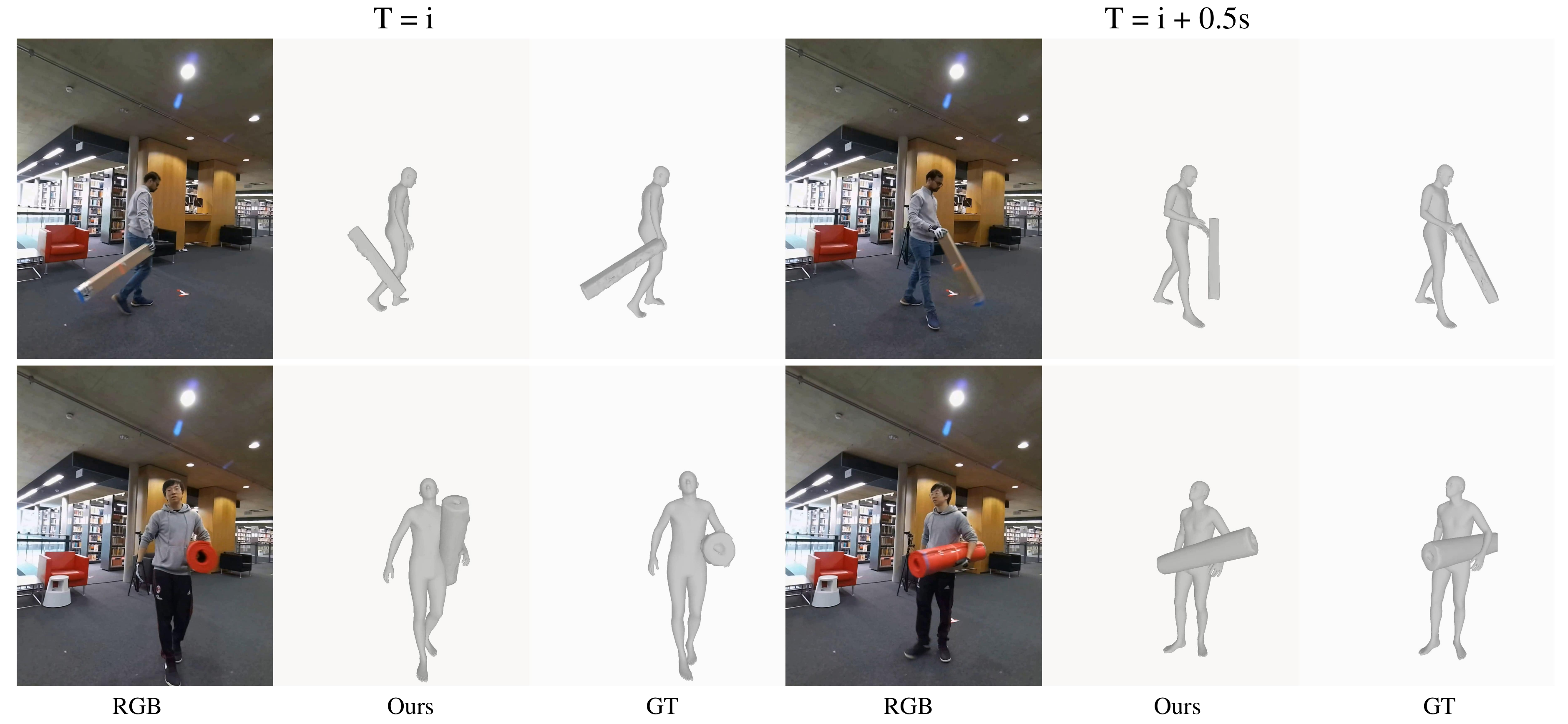}
    \caption{\textbf{Failure cases.} We show reconstruction failures for two elongated objects (top and bottom rows). Each case compares our predicted meshes against the GT at two timestamps separated by 0.5s.}
  \label{fig:failure_cases}
\end{figure}

\paragraph{Degradation in 6-DoF Orientation Tracking.}
As visualized in Fig.~\ref{fig:failure_cases}, THO occasionally struggles to accurately recover the full 6-DoF poses of objects with extreme aspect ratios, even when their global translation trajectories are successfully maintained over time.

The root cause lies in the severe projection ambiguity inherent to such geometries. For elongated shapes, marginal rotational shifts in 3D space translate into wildly disparate 2D visual footprints. When our projection-based 3D Vertex Encoder samples features from these fluctuating 2D projections, it produces a highly volatile and inconsistent sequence of object tokens. Consequently, the Temporal Interact-Aware Transformer (TIAT) is unable to establish reliable cross-frame kinematic correlations. Rather than enforcing temporal continuity, the temporal constraints effectively break down under this severe visual inconsistency, leaving the network vulnerable to isolated and geometrically inaccurate pose predictions.

\paragraph{Potential Solutions.}
To address this architectural vulnerability, future iterations of our framework could incorporate rotation-equivariant visual representations~\cite{fuchs2020se,deng2021vector}. By explicitly preserving 3D rotational symmetries within the encoded latent space, the model could maintain feature consistency independent of the camera viewing angle. This robust geometric grounding would provide TIAT with stable spatial cues, ensuring that temporal constraints remain effective even when reconstructing challenging object shapes.